\documentclass[letterpaper, 10 pt, conference]{ieeeconf}  

\IEEEoverridecommandlockouts                              

\usepackage{graphics} 
\usepackage{graphicx}
\usepackage{booktabs}
\usepackage{amsmath,amssymb,amsfonts}
\usepackage{textcomp}
\usepackage{xcolor}
\usepackage{multirow}
\usepackage{multicol}
\usepackage{subcaption}
\usepackage{amsmath} 
\DeclareMathOperator*{\argmax}{argmax}
\usepackage{amssymb}  
\usepackage{algorithm}
\usepackage[noend]{algpseudocode}
\usepackage{lipsum}
\usepackage{gensymb}

\usepackage{amsthm}
\usepackage{float}
\let\labelindent\relax
\usepackage[inline]{enumitem}
\usepackage[breaklinks,colorlinks]{hyperref}

\title{\LARGE \bf
A Parameter-Efficient Tuning Framework for Language-guided Object Grounding and Robot Grasping
}

\author{Houjian Yu$^{1}$, Mingen Li$^{1}$, Alireza Rezazadeh$^{1}$, Yang Yang$^{2}$, and Changhyun Choi$^{1}$
\thanks{*This work was supported in part by the Sony Research Award Program and NSF Award 2143730.}
\thanks{$^{1}$ The authors are with the Department of Electrical and Computer Engineering, Univ. of Minnesota, Minneapolis, USA
        {\tt\small \{yu000487, li002852, rezaz003, cchoi\}@umn.edu}}
\thanks{$^{2}$ Yang Yang is with the Department of Computer Science and Engineering, Univ. of Minnesota, Minneapolis, USA
        {\tt\small yang5276@umn.edu}}
}

\begin{document}

\maketitle
\thispagestyle{empty}
\pagestyle{empty}

\begin{abstract}
The language-guided robot grasping task requires a robot agent to integrate multimodal information from both visual and linguistic inputs to predict actions for target-driven grasping. While recent approaches utilizing Multimodal Large Language Models (MLLMs) have shown promising results, their extensive computation and data demands limit the feasibility of local deployment and customization. To address this, we propose a novel CLIP-based \cite{radford2021learning} multimodal parameter-efficient tuning (PET) framework designed for three language-guided object grounding and grasping tasks: (1) Referring Expression Segmentation (RES), (2) Referring Grasp Synthesis (RGS), and (3) Referring Grasp Affordance (RGA). Our approach introduces two key innovations: a bi-directional vision-language adapter that aligns multimodal inputs for pixel-level language understanding and a depth fusion branch that incorporates geometric cues to facilitate robot grasping predictions. Experiment results demonstrate superior performance in the RES object grounding task compared with existing CLIP-based full-model tuning or PET approaches. In the RGS and RGA tasks, our model not only effectively interprets object attributes based on simple language descriptions but also shows strong potential for comprehending complex spatial reasoning scenarios, such as multiple identical objects present in the workspace. Project page: \url{https://z.umn.edu/etog-etrg}

\end{abstract}

\section{Introduction}
\begin{figure}[t]
\centering
\includegraphics[width=\linewidth]{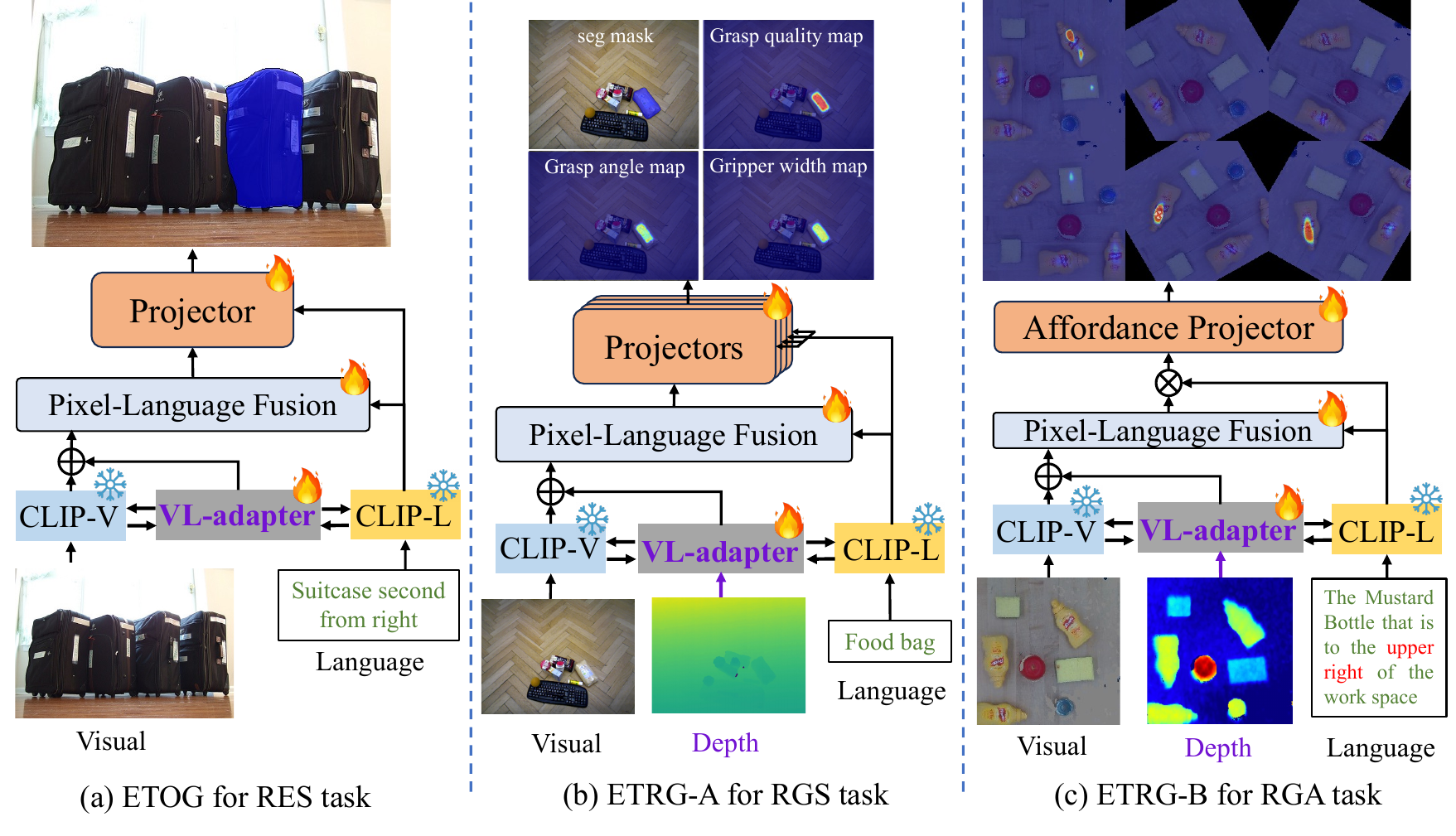}
\vspace{-5mm}
\caption{\textbf{Efficient-Tuning pipeline for language-guided object grounding and robot grasping tasks.} We propose Efficient-Tuning Object Grounding (ETOG) for RES ask, Efficient-Tuning Robot Grasping type-A (ETRG-A) for RGS task, and type-B (ETRG-B) for RGA task. Our framework with minor modifications is able to solve the three tasks. \textit{Zoom-in for more details.}}
\label{task_pipeline:fig}
\vspace{-7mm}
\end{figure}

Target-oriented grasping is a critical yet challenging task in robot manipulation, requiring both precise object identification and high grasping accuracy to retrieve the correct target. To address this, researchers have developed image-driven methods \cite{yu2023iosg, xu2021efficient, lou2021collision, lou2023adversarial, danielczuk2019mechanical}, where the robot receives a reference image of the target object, localizes it using object recognition or matching models \cite{oquab2023dinov2, chen2021exploring, yu2022self, 9382336}, and predicts grasp poses based on detection bounding boxes or segmentation masks. However, object grounding performance degrades when the reference image differs from the target in the workspace due to factors like lighting, viewpoint, or occlusion. 
Another approach focuses on language-guided robot grasping \cite{tziafas2023language, yang2024attribute, ahn2022visually, xu2023joint, yang2021attribute}. While language inputs offer greater flexibility than images, they can introduce ambiguity if not carefully articulated. For instance, as shown in Fig. \ref{task_pipeline:fig}c, if the input is ``mustard bottle" and multiple instances are present, the robot may struggle to grasp the correct one. More detailed linguistic descriptions (e.g., ``The Mustard Bottle that is to the upper right of the workspace" in Fig. \ref{task_pipeline:fig}c) reduce ambiguity but increase the complexity of vision-language comprehension. Previous works primarily focus on interpreting simplified language expressions \cite{yang2024attribute}, using basic colors and shapes \cite{ahn2022visually}, or selecting a distinct target without duplicates \cite{yang2024attribute,xu2023joint}. To address these limitations, an ideal language-guided grasping pipeline should: (1) handle open-vocabulary inputs, (2) interpret complex language descriptions to resolve object ambiguities, and (3) accurately localize and grasp the target with high success rates.

Recent robot grasping studies using Large Language Models (LLMs) or Multimodal Large Language Models (MLLMs) have demonstrated strong object grounding and grasp detection performance \cite{vuong2024language, nguyen2024language, huang2024manipvqa, qian2024thinkgrasp, jin2024reasoning}. However, full-model training or fine-tuning is computationally expensive and may suffer from catastrophic forgetting \cite{kirkpatrick2017overcoming}, limiting the feasibility of local deployment. In this paper, we propose a parameter-efficient tuning (PET) approach~\cite{xu2023bridging, wangbarleria} for object grounding and robot grasping,  enabling competitive performance by incorporate a lightweight module into a frozen pre-trained model. 
%
Specifically, we propose a CLIP-based PET framework for object grounding and robot grasping. As shown in Fig. \ref{task_pipeline:fig}, our framework, with minor modifications on the task projector, addresses three challenging vision-language tasks: Referring Expression Segmentation (RES) \cite{hu2016segmentation}, Referring Grasp Synthesis (RGS) \cite{tziafas2023language, vuong2024language}, and Referring Grasp Affordance (RGA) (see Section \ref{prob_form}) for which the framework processes visual images and complex linguistic descriptions to predict object segmentation masks, grasp-related maps, and grasp affordance maps, respectively.

Our primary contributions are as follows: (1) We propose a bi-directional vision-language adapter that fuses features from multiple modalities and enhances the cross-modal interaction for pixel-level language understanding. By fine-tuning the lightweight adapters (only $0.8\%$ to $2.0\%$ of the frozen CLIP backbone parameters) along with the task decoder, we achieve performance comparable to full-model fine-tuning baselines.
(2) We introduce a depth fusion branch that integrates geometric cues with vision and language features, improving 4-DoF grasp pose predictions.
(3) We present \textbf{E}fficient-\textbf{T}uning for \textbf{O}bject \textbf{G}rounding (\textbf{ETOG}) and \textbf{E}fficient-\textbf{T}uning for \textbf{R}obot \textbf{G}rasping type-A (\textbf{ETRG-A}) and type-B (\textbf{ETRG-B}) models for three key vision-language tasks, with minor modifications to our framework. Extensive evaluations demonstrate the effectiveness of our design in handling open-vocabulary language inputs and complex spatial reasoning.
Each model is trained end-to-end, with the only prior knowledge being the pre-trained CLIP model~\cite{radford2021learning}.

\section{Related Works}

\subsection{Language-guided Object Grounding}
The object grounding task aims to localize the target instance described by a language expression. Based on the output format (bounding boxes or segmentation masks), it can be categorized as referring expression comprehension (REC) or referring expression segmentation (RES). Our proposed method focuses on pixel-level vision-language alignment and mask predictions, primarily designed for the RES task. CLIP-based RES methods leverage the strong text-image alignment of the CLIP model \cite{radford2021learning}, typically combining the CLIP backbone with a multimodal decoder for cross-modal interactions \cite{xu2023bridging, wangbarleria, sun2023language, wang2022cris, yan2023eavl}. Most of these methods use full-model fine-tuning and are limited to the RES task. In contrast, our approach emphasizes parameter-efficient tuning and extends object grounding into robotics tasks.

\subsection{Language-guided Robot Grasping}
Recent advances in multimodal learning have enabled robots to understand human language instructions and generate accurate grasp pose \cite{tziafas2023language, yang2024attribute, vuong2024language, huang2024manipvqa, jin2024reasoning, lu2023vl, tang2023task}. Language-guided grasping typically involves two stages: first, obtaining a target object's bounding box or segmentation mask, then feeding it into a grasping pipeline \cite{lu2023vl, cheang2022learning, sun2023language}. Some researchers have explored end-to-end approaches for generating grasp poses in Referring Grasp Synthesis (RGS) \cite{tziafas2023language, tang2023task}, while others have focused on predicting grasp poses via affordance maps in Referring Grasp Affordance (RGA) \cite{yang2024attribute, shridhar2022cliport}, bypassing the need for object bounding boxes or segmentation masks. Our work addresses both tasks simultaneously within an efficient-tuning framework.

\subsection{Parameter-Efficient Tuning}
Traditional fine-tuning updates all model parameters to adapt to a new task, but as models grow larger, this becomes computationally expensive. Parameter-efficient tuning (PET) methods—such as adapters \cite{houlsby2019parameter, chen2022adaptformer, jie2022convolutional}, prefix-tuning \cite{li2021prefix}, LoRA \cite{hu2021lora}, and prompt-tuning \cite{zhou2022learning}—address this by updating only a small subset of parameters or adding task-specific parameters, while keeping most pre-trained parameters unchanged. These methods reduce memory usage, accelerate training, and facilitate efficient transfer learning, making it easier for roboticists to adapt large pre-trained foundation models to new tasks with lower computational demands. Recent works \cite{xu2023bridging, wangbarleria} focus on object grounding, while our approach combines both object grounding and grasping tasks. Additionally, we introduce a novel method to integrate depth information into the RGB-only pipeline, improving robot grasp predictions.

\section{Methods}
In this section, we propose a parameter-efficient tuning (PET) framework for language-guided object grounding and robot grasping.

\subsection{Problem Formulation} \label{prob_form}

As shown in Fig. \ref{task_pipeline:fig}, we provide detailed formulations for the following three tasks. We use $I \in \mathbb{R}^{H\times W\times 3}$ to represent image input where $(H, W)$ denote the image resolution, $T \in \mathbb{R}^L$ for a language expression, and $D \in \mathbb{R}^{H\times W}$ for a depth image. Our pipeline takes visual images and language descriptions as inputs and outputs task specific predictions as follows.
%
\textbf{RES Task:} We follow the classical RES definition \cite{hu2016segmentation} for this task. Given an image $I$ and a language expression $T$ describing specific parts of the image, the model is tasked to segment the corresponding area(s) of the language description obtaining a segmentation mask $M \in \{0,1\}^{H \times W}$. We refer this task as object grounding.
\begin{figure*}[t]
\begin{center}
    \includegraphics[width=0.85\linewidth]{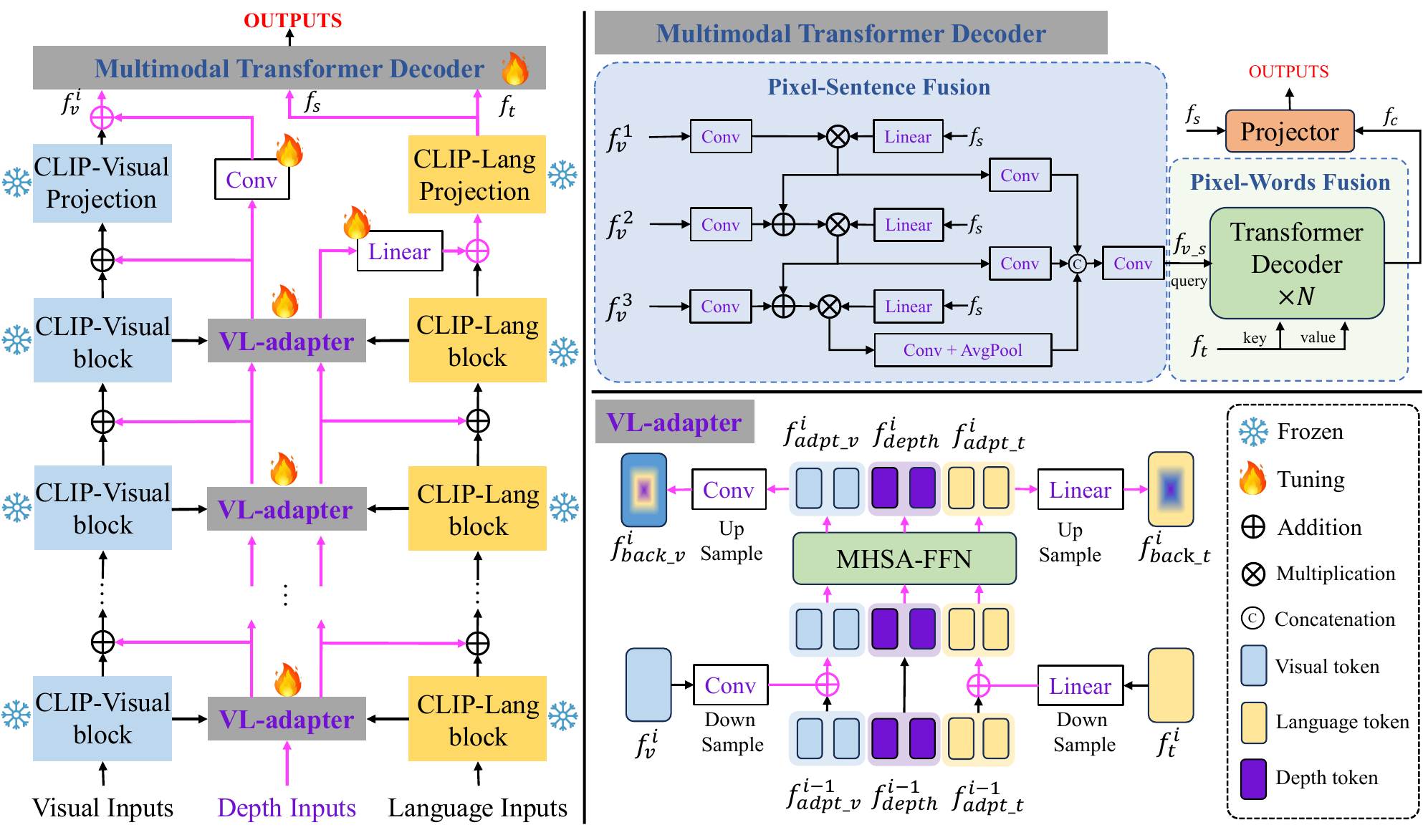}
\end{center}
\vspace{-4mm}
\caption{Our model takes visual, language and depth (optional, depending on tasks) as inputs and outputs the task predictions. Our VL-adapter fuses the multimodal features extracted from CLIP at different stages and previous VL-adapter layers. The Pixel-Sentence Fusion module combines multi-scale visual features with a sentence-level global representation, while the Pixel-Words Fusion module focuses on word-level token understanding and eventually generates the segmentation masks.}
\label{sys:fig}
\vspace{-5mm}
\end{figure*}
%
\textbf{RGS Task:} Given an RGB image $I$, a depth image $D$ and a language expression $T$ referring a unique object in the image, the goal is to predict a segmentation mask of the target object on which 4-DoF grasp poses $G=\{x, y, \theta, l\}$ (a.k.a grasp rectangles) are hypothesized. Here, $(x, y)$ denotes the gripper center location at the image coordinate, $\theta$ indicates the gripper rotation at the camera frame, and $l$ indicates the gripper open width. Following \cite{tziafas2023language}, the model predicts three maps to recover the gripper pose $G$. More specifically, $(x, y)$ is determined by the max value's pixel coordinate in grasp quality map $Q$. The rotation angle is provided by $\theta=\Theta(x,y)$ where $\Theta$ is the grasp angle map. Similarly the gripper open width map $P$ produces $l=P(x,y)$.
%
\textbf{RGA Task:} Given an RGB image $I$, a depth image $D$, and a language expression $T$ referring a target object in the scene, the goal is to grasp the target object based on the predicted pixel-wise grasp affordance maps with $N$ different rotation angles $Q_g \in \mathbb{R}^{H \times W \times N}$ \cite{yang2024attribute}. We also employ a 4-DoF grasp pose $G=\{x^*,y^*,\theta^*,z\}$ to represent the grasping motion, where $(x^*,y^*, \theta^*)=\argmax_{(x, y, \theta)} \,Q_g(x,y,\theta)$ and $z=D(x^*,y^*)$.

\subsection{Multimodal Feature Early Fusion}

The framework details are shown in Fig. \ref{sys:fig}. The model inputs consist of RGB images $I$, language expressions $T$, and optional inputs from a different modality (e.g. depth images $D$) depending on the task type. The frozen CLIP image encoder and text encoder extract $i$-th early-stage visual and language intermediate features $f_v^i \in \mathbb{R}^{H_i \times W_i \times C_i}$ and $f_t^i \in \mathbb{R}^{L \times C_t}$, where $H_i, W_i, C_i$ are the height, width and number of channels of $f_v^i$ respectively, $L$ is the language token length, $C_t$ is the number of channels of the language token, $i \in \{1, 2, ..., N\}$, and $N$ denotes the number of CLIP vision ResNet \cite{he2016deep} or vision Transformer (ViT) \cite{dosovitskiy2020image} blocks based on the choice of CLIP backbones. 

To enhance cross-modal feature interaction between the originally separated vision and language encoders, we propose our novel VL-adapter to execute vision-language feature early fusion. Before feeding the vision $f_v^{i}$ and language features $f_t^{i}$ from previous layers to the next CLIP visual block $\phi_i$ and language block $\psi_i$, $i\in\{2,..., N\}$, our VL-adapter takes them as input and outputs the refined multimodal features to the corresponding CLIP branch.

In VL-adapter, we first adjust the $f_v^{i}$ and $f_t^{i}$ to have the same feature dimension by passing each of them to a convolution layer and a linear layer, respectively. Then, the downsampled visual and linguistic features are added with $f_{adpt_v}^{i-1}$ and $f_{adpt_t}^{i-1}$ passed from the former VL-adapter
$\hat{f}_v^i=\text{Conv}({f}_{v}^i) + f_{adpt\_v}^{i-1}, \hat{f}_t^i=\text{Linear}({f}_{t}^i) + f_{adpt\_t}^{i-1}$
where $\text{Conv}(\cdot)$ and $\text{Linear}(\cdot)$ denote convolution and linear layers, respectively. Such an operation reduces computation complexity meanwhile makes sure that the features obtained from the previous stage are able to get to the deeper layers.

Moreover, we concatenate the visual, linguistic, and the optional input feature (e.g., depth feature) from a different modality. For RES task, it only requires visual and linguistic features. The concatenated tokens are then fed into a transformer layer \cite{vaswani2017attention} using the following equation:
\begin{equation}
\begin{aligned}
& f_{vdt}=\text{Concat}(\text{LN}(\hat{f}_v^i), \text{LN}(f_{depth}^{i-1}), \text{LN}(\hat{f}_t^i)) \\
& f_{vdt}^{\prime} = \text{MHSA}(f_{vdt}) + f_{vdt} \\
& f_{vdt}^{\prime\prime} = \text{FFN}(f_{vdt}^\prime)+f_{vdt}^\prime \\
&[f_{adpt\_v}^i, f_{depth}^i, f_{adpt\_t}^i]= \text{Split}(f_{vdt}^{\prime\prime})
\end{aligned}
\end{equation}
where Concat$(\cdot)$, LN$(\cdot)$, MHSA$(\cdot)$, FFN$(\cdot)$, and Split$(\cdot)$ represent concatenation operation, layer normalization, multi-head self-attention layer, feed-forward network, and split operation, respectively. In addition, our VL-adapter outputs the multimodal fused features back to CLIP vision and language branches updating $f_v^i$ and $f_t^i$:
$f_v^i=f_{back\_v}^i + f_{v}^{i}$ where $f_{back\_v}^i = \text{Conv}({f}_{adpt\_v}^i)$ and $f_t^i=f_{back\_t}^i + f_{t}^{i}$ where $f_{back\_t}^i = \text{Linear}({f}_{adpt\_t}^i)$.

Unlike the previous in-CLIP design \cite{xu2023bridging} where the fused features did not directly forward outside of CLIP model, we add two stand-alone convolution layer and linear layer (as shown on the left side of Fig. \ref{sys:fig}) to make sure that the VL-adapter fused information is able to engage in feature interaction at deep layers. This simple but effective modification helps the object grounding task (see Section \ref{exp:res}). In this case, CLIP model together with VL-adapter will provide visual features $f_v^i$ at multiple stages, sentence-level feature $f_s \in \mathbb{R}^{C_s}$ and word-level feature $f_t \in \mathbb{R}^{L\times C_t}$ where $C_s$ and $C_t$ are the feature dimensions.

\subsection{Multimodal Transformer Decoder}

\noindent\textbf{Pixel-Sentence Fusion:} We propose a pixel-sentence fusion module to learn sentence-level vision-language features (Fig. \ref{sys:fig}). Given visual features $f_v^i$ extracted at different stages from the CLIP visual branch and the sentence feature $f_s$, we fuse such multi-scale visual features and sentence representations to obtain pixel-sentence multimodal feature $f_{v\_s}$. Element-wise multiplication and addition operations are applied to this fusion module.

\noindent\textbf{Pixel-Words Fusion:} As shown in Fig. \ref{sys:fig}, the pixel-words fusion module takes the pixel-sentence feature $f_{v\_s}$ and the word-level feature from CLIP model as inputs and executes cross-modal feature interactions in a transformer architecture. Each transformer decoder layer consists of a multi-head self-attention layer, a multi-head cross-attention layer, and a feed-forward network layer. To begin with, the model computes self-attention on the pixel-sentence feature $f_{v\_s}$:
$f_{v}^{\prime}= \text{LN}(\text{MHSA}(f_{v\_s})+f_{v\_s})$.
We apply post layer normalization through this transformer decoder. Then, we adopt cross-attention to the evolved visual feature $f_{v}^{\prime}$:
$f_{v}^{\prime}= \text{LN}(\text{MHCA}(f_{v}^{\prime}, f_t)+f_{v}^{\prime}), f_c = \text{LN}(\text{FFN}(f_{v}^{\prime})+f_{v}^{\prime})$
where MHCA$(\cdot)$ indicates cross-modal attention. Through such a process, we obtain the multimodal feature $f_c$ which combines multi-scale information from different modalities.


\subsection{Projectors and Training Objectives} Projectors in ETOG and ETRG-A consist of standard convolution and linear layers. They transfer $f_c$ and $f_s$ into $F_c \in \mathbb{R}^{N\times D}$ where $N=\frac{H}{4}\times \frac{W}{4}$ and $F_s \in \mathbb{R}^{C}$ where $C=K\times K \times D + 1$, respectively. $K$ denotes a customized kernel size and $F_s$ serves as a convolution kernel to operate computation with $F_c$ feature map. We employ a text-to-pixel contrastive loss following \cite{wang2022cris} to train the object segmentation projectors. The training loss $L_{\text{tp}}$ can be formulated as follows:
$L_{\text {tp}}\left(F_t, F_c\right) =\frac{1}{|\mathcal{P} \cup \mathcal{N}|} \sum_{i \in \mathcal{P} \cup \mathcal{N}} L_{\text {contrast}}^i\left(F_s, F_c^i\right)$
where $\mathcal{P}$ and $\mathcal{N}$ denotes the class of 1 and 0, respectively, and  
$L_{\text {contrast}}^i\left(F_s, F_c^i\right) = -\log \left(\sigma\left(F_s \cdot F_c^i\right)\right)$ if $i \in \mathcal{P}$ or 
$-\log \left(1-\sigma\left(F_s \cdot F_c^i\right)\right)$ if $i \in \mathcal{N}$
where $\sigma$ represents the sigmoid function.

Besides $L_{\text{tp}}$, we duplicate the segmentation head three times to supervise the grasp quality maps, grasp angle maps, and gripper open width maps predictions using smooth L1 loss for the RGS task.

For the RGA task, ETRG-B uses the fully convolutional network to decode the multimodal feature for the affordance maps $Q_g$ prediction. To represent grasp pose with different angles, we rotate the vision-language feature $F_a$, derived from element-wise multiplication of $f_s$ and $f_c$, by $N=6$ (i.e., multiples of $\theta=30\degree$) orientations predicting pixel-wise scores of horizontal grasps within the rotated maps. 
Since this task does not require object grounding but directly chooses the pixel with the highest score among all the $N$ maps, we further employ the motion loss $\mathcal{L}_{grasp}$ from our previous work \cite{yang2024attribute}.

\section{Experiments}

\begin{table*}[t]
    \caption{Performance comparison of different SOTA CLIP-based models on RefCOCO related datasets using mIoU metric. Backbone: visual encoder architecture. U: the UMD test split for RefCOCO(g). Params: total number of parameters to be tuned.}
    \vspace{-1mm}
    \centering
    \footnotesize
    \begin{tabular}{l|l|l|c|c|c|c|c|c|c|c|c|c}
    \hline
        \multirow{2}{*}{Baselines} & \multirow{2}{*}{Conference} & \multirow{2}{*}{Backbone} & \multicolumn{3}{c|}{RefCOCO} & \multicolumn{3}{c|}{RefCOCO+} & \multicolumn{2}{c|}{RefCOCO(g)} & \multirow{2}{*}{AVG} & \multirow{2}{*}{Params} \\
        \cline{4-11}
        & & & val & testA & testB & val & testA & testB & val(U) & test(U) & & \\
        \hline
        CLIPORT-semantic \cite{shridhar2022cliport} & CORL21 & CLIP-R50 & 57.02 & 61.56 & 49.16 & 43.25 & 51.29 & 31.99 & 42.20 & 43.39 & 47.48 & 51.82M \\
        CRIS \cite{wang2022cris} & CVPR22 & CLIP-R50 & 69.52 & 72.72 & 64.70 & 61.39 & 67.10 & 52.48 & 59.35 & 59.39 & 63.33 & 146.85M \\
        DyCRIS-M \cite{sun2023language} & IROS23 & CLIP-R50 & 70.54 & 73.13 & 65.63 & 62.28 & 67.61 & 53.25 & 60.80 & 60.87 & 64.26 & 146.85M \\
        ETRIS \cite{xu2023bridging} & ICCV23 &CLIP-R50 & 70.39 & 73.11 & 66.38 & 60.47 & 67.11 & 50.73 & 59.71 & 59.95 & 63.48 & 25.66M \\
        \textbf{ETOG (Ours)} & ICRA25 & CLIP-R50 & \textbf{72.31} & \textbf{75.49} & \textbf{66.62} & \textbf{63.26} & \textbf{69.46} & \textbf{53.27} & \textbf{60.80} & \textbf{61.98} & \textbf{65.40} & 32.93M \\
        \hline 
        CRIS & CVPR22 & CLIP-R101 & 70.47 & 73.18 & 66.10 & 62.27 & 68.08 & 53.68 & 59.87 & 60.36 & 64.25 & 161.25M \\
        ETRIS & ICCV23 & CLIP-R101 & 71.06 & 74.11 & 66.66 & 62.23 & 68.51 & 52.79 & 60.28 & 60.42 & 64.51 & 25.92M \\
        EAVL \cite{yan2023eavl} & Arxiv23 & CLIP-R101 & 70.65 & 73.81 & 66.62 & 63.42 & 68.26 & 54.71 & 60.93 & 61.24 & 64.96 & - \\
        \textbf{ETOG (Ours)} & ICRA25 & CLIP-R101 & \textbf{73.37} & \textbf{76.16} & \textbf{68.54} & \textbf{64.70} & \textbf{70.86} & \textbf{54.94} & \textbf{63.06} & \textbf{63.66} & \textbf{66.91} & 30.57M \\
        \hline
        ETRIS & ICCV23 & CLIP-ViT-B & 70.51 & 73.51 & 66.63 & 60.1 & 66.89 & 50.17 & 59.82 & 59.91 & 63.44 & 25.37M \\
        EAVL & Arxiv23 & CLIP-ViT-B & 71.74 & 74.38 & 67.24 & 64.05 & 68.41 & 56.82 & 61.87 & 62.21 & 65.84 & - \\
        BARLERIA \cite{wangbarleria} & ICLR24 & CLIP-ViT-B & 72.40 & 75.90 & 68.30 & 65.00 & 70.80 & 56.90 & 63.40 & 63.80 & 67.06 & - \\
        \textbf{ETOG (Ours)} & ICRA25 & CLIP-ViT-B & \textbf{73.37} & \textbf{76.90} & \textbf{69.34} & \textbf{65.95} & \textbf{71.47} & \textbf{56.94} & \textbf{63.75} & \textbf{64.63} & \textbf{67.79} & 29.89M \\
        \hline
    \end{tabular}
    \vspace{2mm}
    \label{tab:clip_comparison}

    \centering
    \caption{Comparison on full model fine-tuning RES methods with RefCOCO related datasets using oIoU metric. Backbone: visual encoder architecture. U: the UMD test split for RefCOCO(g) dataset.}
    \vspace{-1mm}
    \footnotesize
    \begin{tabular}{l|l|l|c|c|c|c|c|c|c|c|c}
        \hline
        \multirow{2}{*}{Baselines} & \multirow{2}{*}{Conference} & \multirow{2}{*}{Backbone} & \multicolumn{3}{c|}{RefCOCO} & \multicolumn{3}{c|}{RefCOCO+} & \multicolumn{2}{c|}{RefCOCO(g)} & \multirow{2}{*}{AVG} \\
        \cline{4-11}
        & & & val & testA & testB & val & testA & testB & val(U) & test(U) & \\
        \hline
        PCAN \cite{chen2022position} & Arxiv22 & ResNet-50 & 69.51 & 71.64 & 64.18 & 58.25 & 63.68 & 48.89 & 59.98 & 60.80 & 62.12 \\
        VLT \cite{ding2021vision} & ICCV21 & DarkNet-53 & 65.65 & 68.29 & 62.73 & 55.50 & 59.20 & 49.36 & 52.99 & 56.65 & 58.80 \\
        MATTNET \cite{yu2018mattnet} & CVPR18 & ResNet-101 & 56.51 & 62.37 & 51.70 & 46.67 & 52.39 & 40.08 & 47.64 & 48.61 & 50.75 \\
        CGAN \cite{luo2020cascade} & ACM20 & ResNet-101 & 64.86 & 68.04 & 62.07 & 51.03 & 55.51 & 44.06 & 51.01 & 51.69 & 56.03 \\
        DMMI \cite{hu2023beyond} & ICCV23 & ResNet-101 & 68.56 & 71.25 & 63.16 & 57.90 & 62.31 & 50.27 & 59.02 & 59.24 & 61.46 \\
        ReSTR \cite{kim2022restr} & CVPR22 & ViT-B & 67.22 & 69.30 & 64.45 & 55.78 & 60.44 & 48.27 & 54.48 & - & 59.99 \\
        \hline
        ETOG (Ours) & ICRA25 & ResNet50 & 70.11 & 73.97 & 63.76 & 58.63 & 65.66 & 47.86 & 57.35 & 59.87 & 62.15 \\
        ETOG (Ours) & ICRA25 & ResNet101 & 71.11 & 74.30 & 65.61 & 60.00 & 67.27 & 49.73 & 60.09 & 61.23 & 63.67 \\
        \textbf{ETOG (Ours)} & ICRA25 & ViT-B & \textbf{71.35} & \textbf{76.14} & \textbf{66.73} & \textbf{62.33} & \textbf{68.52} & \textbf{51.90} & \textbf{61.12} & \textbf{62.85} & \textbf{65.12} \\
        \hline
    \end{tabular}
    \vspace{-5mm}
    \label{tab:res_comparison}
\end{table*}

In this section, we conduct language-guided object grounding and robot grasping experiments to evaluate the proposed method. The experiment goals are to verify if: 1) our Efficient-Tuning for Object Grounding (\textbf{ETOG}) generic model is able to correctly ground the target object, 2) the proposed depth fusion branch in Efficient-Tuning for Robot Grasping (\textbf{ETRG-A}) can contribute to more accurate grasping prediction, and 3) the \textbf{ETRG-B} model manages to understand object-attribute descriptions and achieve high grasping success rate with more complex spatial relationship language inputs. All our models are trained and tested on a workstation with a single NVIDIA RTX 2080 Ti.

\subsection{Referring Expression Segmentation for Object Grounding} \label{exp:res}

\noindent\textbf{RES Datasets:} we evaluate our \textbf{ETOG} generic method on three standard RES benchmark datasets: RefCOCO \cite{kazemzadeh2014referitgame}, RefCOCO+ \cite{kazemzadeh2014referitgame}, and RefCOCOg \cite{mao2016generation, nagaraja2016modeling} that include $19,994$, $19,992$, and $26,771$ RGB images, with $142,209$, $141,564$, and $104,560$ annotated language expressions, referring to $50,000$, $49,856$, and $54,822$ annotated objects, respectively. RefCOCO+ is designed to be more challenging by excluding the absolute spatial location description, and RefCOCOg dataset has an average of 8.4 words length in the language descriptions

\noindent\textbf{Evaluation Metrics:} We adopt the mean intersection of union (mIoU), overall intersection of union (oIoU), and $Prec@X$ to verify the object grounding effectiveness. 
The $Prec@X$ evaluates the percentage of test results with an IoU value higher than $X \in \{0.5, 0.7, 0.9\}$.

\noindent\textbf{Implementation Details:} We train ETOG for 50 epochs with a batch size 16 using the Adam optimizer with an initial learning rate of 5e-5. The learning rate is decreased with a decay factor of 0.1 at the $35^{th}$ epoch.

\begin{figure}[ht]
\begin{center}
    \includegraphics[width=\linewidth]{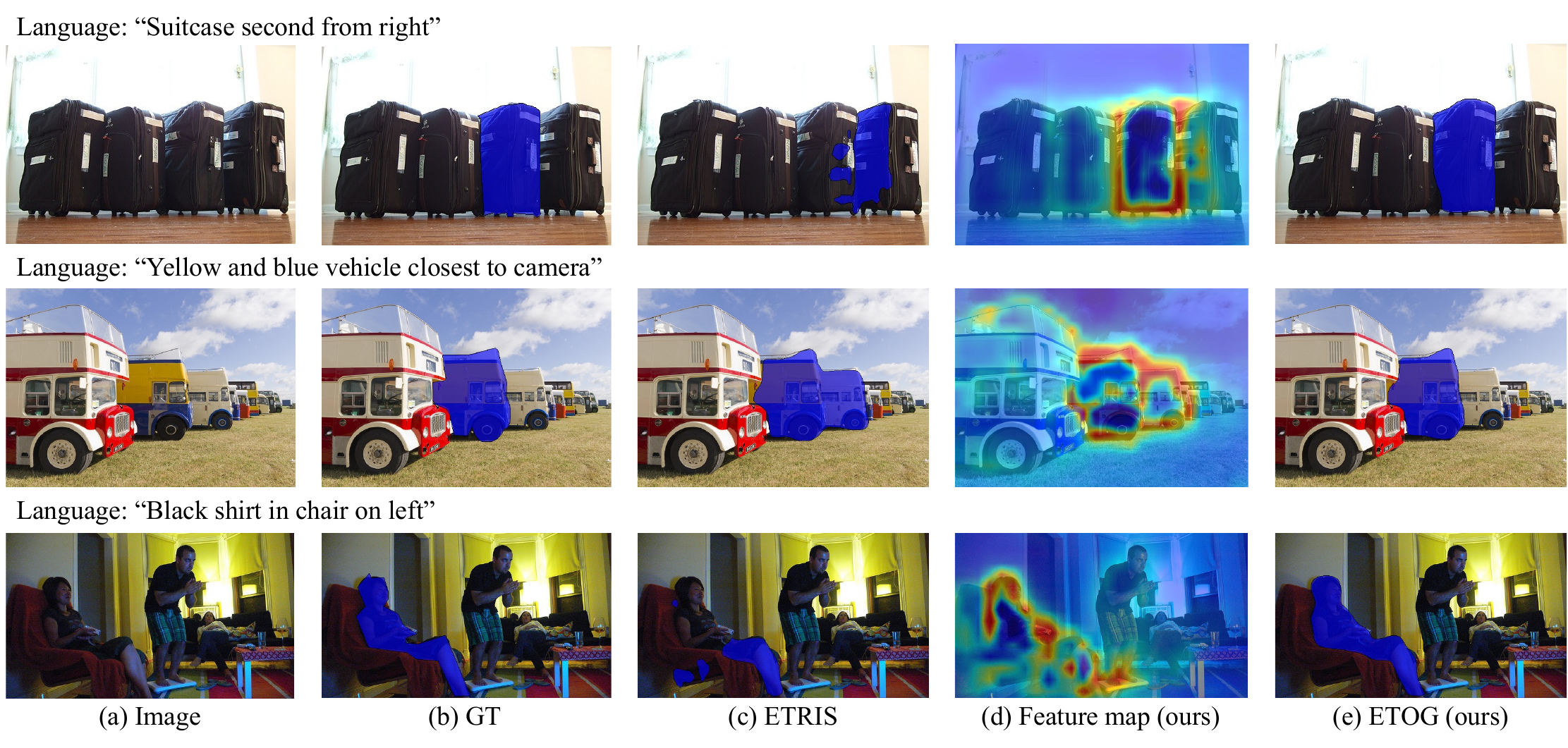}
\end{center}
\vspace{-4mm}
\caption{\textbf{Qualitative results on model prediction and feature visualization.} \textit{Zoom-in for more details.}}
\label{etog_vis:fig}
\vspace{-5mm}
\end{figure}

\noindent\textbf{ETOG Comparison with RES Baselines:} In Table \ref{tab:clip_comparison}, we compare our ETOG with the state-of-the-art CLIP-based RES methods on RefCOCO datasets using mIoU metric. ETOG outperforms all baseline methods across different backbone architectures. CLIPORT-semantic, a variant of CLIPORT \cite{shridhar2022cliport}, excludes the depth processing branch and uses simple element-wise multiplication to fuse vision and language features. The comparison between CLIPORT-semantic and our method highlights the effectiveness of our VL-adapter. In particular, with the CLIP \cite{radford2021learning} ResNet-101 \cite{he2016deep} backbone, our method--using only $1.9$M tunable parameters--achieves a $2.66\%$ and $1.95\%$ performance gain across 8 evaluation tasks compared to CRIS \cite{wang2022cris} and EAVL \cite{yan2023eavl} that are fully CLIP fine-tuned methods. Additionally, ETOG with $1.21$M tunable ViT-B-16 backbone parameters shows a $4.35\%$ and $0.73\%$ performance improvement compared to the latest parameter-efficient tuning approaches, ETRIS \cite{xu2023bridging} and BARLERIA \cite{wangbarleria}, respectively. Table \ref{tab:res_comparison} further compares our model with full fine-tuned RES models, showing that ETOG achieves competitive object grounding performance even against non-CLIP-based approaches. Fig. \ref{etog_vis:fig} provides qualitative results, including model predictions and feature map visualizations, with the fourth column showing the attention regions of the well-fused multimodal feature $f_c$. 
\noindent\textbf{ETOG Ablation Study.} An ablation study of our VL-adapter can be found in Table \ref{tab:res_ablation}. To make a fair comparison, we only replace our VL-adapter with the corresponding baselines' vision-language fusion modules. It is noticeable that our proposed VL-adapter which combines the vision language tokens and execute multimodal self-attention obtains better mIoU performance compared with previous SOTA vision-language fusion solutions such as parameter-efficient tuning ETRIS's Bridger \cite{xu2023bridging} and LAVT's Pixel-Words Attention Module (PWAM) \cite{yang2022lavt}. We also notice that simply adding forward layers (the stand-alone convolution and linear layers shown in Fig. \ref{sys:fig}) boosts the object grounding performance. In this case, the updated fusion features are forwarded outside of CLIP model and participate the multimodal feature interaction at the deeper layers.

\subsection{Referring Grasp Synthesis with Depth Fusion}

\noindent\textbf{RGS Dataset:} We evaluate \textbf{ETRG-A} on OCID-VLG \cite{tziafas2023language} dataset. OCID-VLG dataset presents a collection of indoor tabletop objects scenes with human annotated image-text-masks-grasp box tuples for referring object grounding and grasp pose prediction tasks. This dataset includes $1,763$ scenes with 58 unique object instances and over $89.6k$ language expressions which contain multiple types of instance-level descriptions.

\begin{table}[t]
    \centering
    \caption{\textbf{ETOG} VL-adapter ablation study. Our proposed VL-adapter with the Forward Layers (FL) achieves the best performance. The following baselines are evaluated on RefCOCO validation dataset with CLIP-R50 backbone. $(*)$ denotes that we re-implement the baseline with its default setup.}
    \begin{tabular}{l|l|c|c|c}
        \hline
        Baselines & mIoU & Pr@50 & Pr@70 & Pr@90 \\
        \hline
        ETRIS$^*$ \cite{xu2023bridging} & 69.60 & 81.87 & 69.53 & 16.00 \\
        LAVT-PWAM \cite{yang2022lavt} \ & 71.25 & 82.21 & 70.83 & 24.66 \\
        VL-adapter (Ours) & 71.74 & 82.95 & 71.49 & 24.83 \\
        \textbf{VL-adapter+FL (Ours)} & \textbf{72.31} & \textbf{83.22} & \textbf{72.95} & \textbf{26.17}
        \\
        \hline
    \end{tabular}
    \vspace{-4mm}
    \label{tab:res_ablation}
\end{table}

\noindent\textbf{Evaluation Metrics:} We report mean intersection of union (mIoU) for the referring object grounding task in the object segmentation format. For object grasping prediction, we apply the Jacquard Index $J@N$ metric \cite{tziafas2023language, depierre2018jacquard} measuring the top-$N$ grasp rectangle candidates that have a rotation angle difference less than $30^{\circ}$ and IoU over 0.25 compared with the ground truth grasp rectangle.

\noindent\textbf{Implementation Details:} We train ETRG-A for 40 epochs with a batch size 11. We adopt the AdamW optimizer with an initial learning rate of $\lambda=0.0001$ and a polynomial learning rate decay. The images are resized to $416 \times 416$ resolution, and the maximum sentence length is set to $20$.

\noindent\textbf{Ablation Study on Depth Fusion Branch:} One primary SOTA baseline we evaluate against is CROG \cite{tziafas2023language}, a CLIP-based RGS method with full model fine-tuned with OCID-VLG dataset. As shown in Table \ref{tab:rgs_comparison}, ETRG-A without depth model (ResNet-50) achieves a slightly better ($+3.96\%$ $J@1$) performance and competitive object grounding (RES) performance compared CROG. After integrating depth inputs into VL-adapter (details in Fig. \ref{sys:fig}) fusing the object geometric information into visual-language branches, our ETRG-A with depth fusion full model attains higher grasping rectangle prediction ($+11.86\%$ $J@1$) accuracy with ResNet-50 backbone and ($+12.18\%$ $J@1$) accuracy with ResNet-101 backbone compared with CROG. 
CROG is marginally better than our models in the RES task, but we attribute this to the simplicity of the OCID-VLG dataset, which primarily features tabletop objects and lacks the scene diversity found in RefCOCO-related datasets. This environment favors full model fine-tuning methods like CROG, potentially leading to over-fitting.
Although CROG has 147M tunable parameters which is 3 times larger than our proposed depth-fusion approach, our ETRA-A model still achieves competitive performance on this RES task.
\begin{table}[t]
    \centering
    \caption{\textbf{ETRG-A} ablation study. Our proposed depth fusion branch improves the grasping rectangle prediction performance.}
    \begin{tabular}{l|l|c|c|c}
        \hline
        \multirow{2}{*}{Baselines} & \multirow{2}{*}{Backbone} & \multicolumn{1}{c|}{RES} & \multicolumn{2}{c}{RGS} 
        \\
        \cline{3-5}
        & & mIoU & J@1 & J@Any \\
        \hline
        CROG & ResNet-50 & \textbf{81.10} & 77.20 & 87.70 \\
        ETRG-A w/o depth (Ours) & ResNet-50 & 79.06 & 81.16 & 89.11 \\
        ETRG-A (Ours) & ResNet-50 & 79.82 & 89.06 & 92.17 \\
        ETRG-A w/o depth (Ours) & ResNet-101 & 79.47 & 82.28 & 91.12 \\
        ETRG-A (Ours) & ResNet-101 & 80.11 & \textbf{89.38} & \textbf{93.49} \\
        \hline
    \end{tabular}
    \vspace{-6mm}
    \label{tab:rgs_comparison}
\end{table}

\subsection{Referring Grasp Affordance for Spatial Reasoning}

The proposed ETRG-B model is trained using self-supervision in the CoppeliaSim simulator~\cite{6696520}. The primary goal of this RGA experiment is to validate the model's spatial reasoning capabilities as a beneficial byproduct.

\noindent\textbf{Data Collection:} We generate scenes with seven objects, randomly sampled from 32 complex objects, primarily from YCB objects \cite{calli2015ycb}. Ground truth labels are generated using language descriptions based on object attributes, such as color, shape, category name, and spatial location. We employ several language templates to describe both the \textbf{Absolute} (target object's relation to the workspace) and \textbf{Relative} (target object's relation to other reference objects) spatial relationships. The robot agent either randomly selects a target object or employs ETRG-B affordance maps for grasping. Object relabeling is applied if an incorrect object is grasped. For model training, we collected more than $20,000$ visual-language-grasp triplets using domain randomization. For testing, $1,600$ test cases were generated across the 32 objects, with pre-selected query language descriptions based on target object's spatial location. All baselines were tested on the same scenes and language expressions.

\noindent\textbf{Implementation Details:} We apply online training during the data collection process. Afterward, the entire dataset is replayed for 40 epochs with a batch size of 32. We adopt the AdamW optimizer, with an initial learning rate of 5e-5 and a polynomial learning rate decay.
\begin{table}[ht]
    \centering
    \caption{\textbf{ETRG-B} grasp success rate in simulation. This experiment is designed for model ablation study on spatial reasoning in simulation. We employ CLIP-R50 backbone.}
    \begin{tabular}{l|c|c|c|c}
        \hline
        \multirow{2}{*}{Baselines} & \multicolumn{2}{c|}{Absolute} & \multicolumn{1}{c|}{Relative} & \multirow{2}{*}{AVG}
        \\
        \cline{2-4}
        & 4-obj & 7-obj & 7-obj \\
        \hline
        ETRG-B w/o depth (Ours) & 87.81 & 86.25 & 86.18 & 86.75 \\
        ETRG-B (Ours) & \textbf{90.75} & \textbf{88.38} & \textbf{86.56} & \textbf{88.56} \\
        \hline
    \end{tabular}
    \vspace{-2mm}
    \label{tab:rga-sim}
\end{table}

\noindent\textbf{Simulation Results:} We evaluate our methods on affordance prediction with spatial reasoning. The experiments include two primary scene types: \textbf{Absolute} and \textbf{Relative}. In each test scene, there are three identical object instances, and spatial language expression are used to specify the target. To increase the difficulty of spatial reasoning, we create test cases with four objects and seven objects. As illustrated in Table \ref{tab:rga-sim}, both the proposed methods, with and without depth fusion, demonstrate strong performance in handling complex spatial reasoning tasks. Our model with depth fusion achieves slightly higher performance compared to the RGB-only method.
\begin{figure}[t]
\begin{center}
    \includegraphics[width=\linewidth]{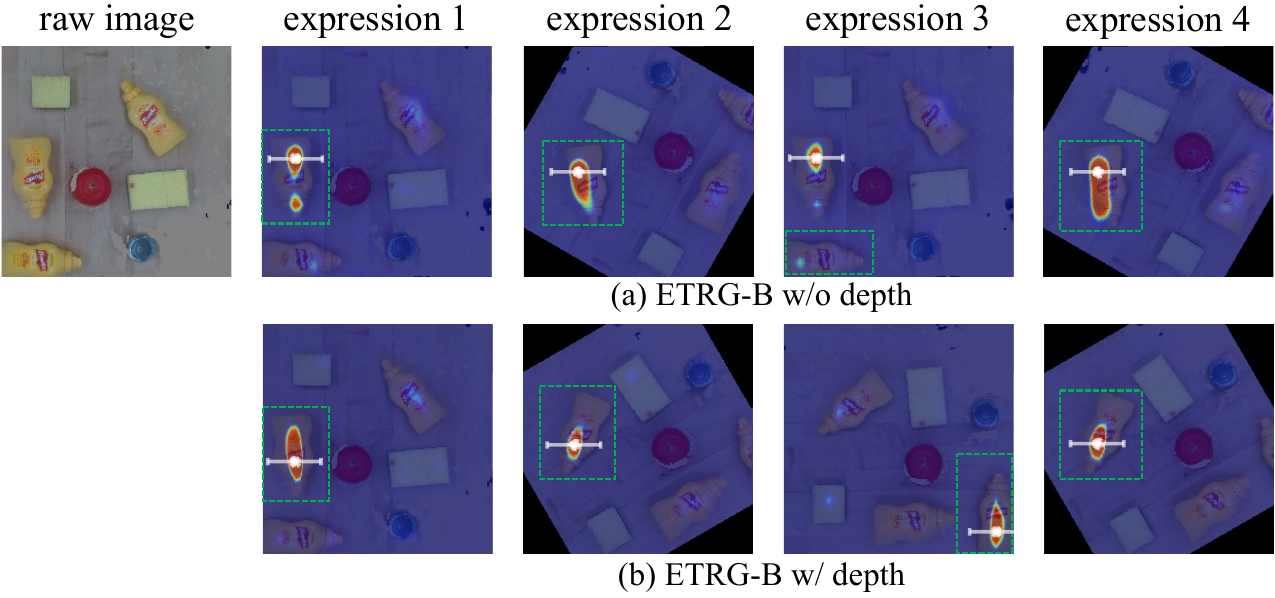}
\end{center}
\vspace{-3mm}
\caption{\textbf{Qualitative results on grasp affordance prediction with ETRG-B.} The green bounding boxes highlight the language-referred target. The language expressions used are: (1) ``The Mustard Bottle that is to the middle left of the workspace", (2) ``Yellow Mustard Bottle that is to the upper right of the workspace", (3) ``lower left Mustard Bottle" and (4) ``Yellow Mustard Bottle that is to the upper right of the apple".}
\label{etrg-b:fig}
\vspace{-1mm}
\end{figure}

\noindent\textbf{Real Robot Results:} We conducted experiments using a Franka Emika Panda robot equipped with a FESTO DHAS soft gripper. We collected over 10 household objects and created scenes with 6 randomly selected objects. Each scene included up to 3 duplicated instances of one category to test spatial reasoning. For each baseline, we report object grounding accuracy, which assesses whether the grasp is on the correct target, and the overall grasp success rate. A detailed visualization of correct mustard bottle identification with feasible grasp poses is shown in Fig. \ref{etrg-b:fig}. Across 40 grasp attempts with different object arrangements (Table \ref{real:tab}), both methods achieve comparable grounding performance, with our model incorporating depth information showing slightly better grasping accuracy.
\begin{table}[t]
\vspace{-1mm}
\caption{\textbf{ETRG-B} Real robot results on RGA spatial reasoning.}
\vspace{-4mm}
\label{real:tab}
\begin{center}
\begin{tabular}{lcc}
\toprule
   Methods & Grounding (\%) & Grasp Succ. (\%) \\
\midrule
ETRG-B w/o depth (Ours) & \textbf{85.0} & 70.0  \\
ETRG-B & \textbf{85.0} & \textbf{75.0} \\
\bottomrule
\end{tabular}
\end{center}
\vspace{-6mm}
\end{table}
\section{Conclusion}
In this work, we presented a novel CLIP-based parameter-efficient tuning framework that can be adapted to a range of tasks, including RES, RGS, and RGA tasks. Our method incorporated a vision-language fusion adapter and a depth-fusion branch, enabling effective open-vocabulary object grounding and robot grasping. We further explored the spatial reasoning capabilities in robot grasping, which the original CLIP \cite{radford2021learning} model does not address. The proposed models demonstrated competitive performance across these challenging vision language and robotics tasks. The current method may under-perform in highly cluttered scene.




{
\bibliography{IEEEexample,IEEEabrv}
\bibliographystyle{IEEEtran}
}

\end{document}